%% file: main.tex
\documentclass{article} 
\usepackage{iclr_templates/iclr2025_conference,times}
\iclrfinalcopy
\input{iclr_templates/math_commands}

\usepackage{hyperref}
\usepackage{url}
\usepackage{microtype}
\usepackage{graphicx}
\usepackage{enumitem}
\usepackage{subfigure}
\usepackage{booktabs} 
\usepackage{wrapfig}
\usepackage{booktabs}
\usepackage{multirow}
\usepackage{adjustbox} 
\usepackage{xcolor,colortbl}
\usepackage{pgfmath}
\usepackage{rotating}
\usepackage{siunitx}
\usepackage{makecell}
\usepackage{algorithm}
\usepackage{algpseudocode}

\title{The High Explosives and Affected Targets (HEAT) Dataset.}

\author{Bryan Kaiser\thanks{Contact: \texttt{bkaiser@lanl.gov}}$\;^{1}$,
   Kyle Hickmann$\;^{1}$,
   Sharmistha Chakrabarti$\;^{1}$,
   Soumi De$\;^{1}$,
   Sourabh Pandit$\;^{1}$,
   \\
   \textbf{David Schodt$\;^{1}$,
   Jesus Pulido$\;^{1}$,
   Divya Banesh$\;^{1}$,
   Christine Sweeney$\;^{1}$}
   \vspace{8pt}
   \\
   \textnormal{$^{1\,}$Los Alamos National Laboratory},   
}
 
\begin{document}

\maketitle

\begin{abstract}
Artificial Intelligence (AI) surrogate models offer a computationally efficient alternative to full-physics simulations, yet no existing datasets are publicly available for training, testing, and validation of machine learning models of the dynamics of high-explosive driven shocks through multiple materials. Shock propagation through materials is a computationally challenging problem because simulations must include material-specific equations of state (EOS) along with descriptions of other physical processes such as plastic deformation, phase change, damage processes, fluid instabilities, and multi-material interactions. Shocks are typically initiated by high-velocity impacts or explosive loading. The latter case necessitates the addition of models of reactive materials to represent high-explosive (HE) detonation.
To address the lack of an expansive dataset for multi-material shock propagation in the AI/ML community, we present the High-Explosives and Affected Targets (HEAT) Dataset. HEAT is a physics-rich collection of two-dimensional, cylindrically symmetric, simulations generated using an Eulerian, multi-material, shock-propagation code developed at Los Alamos National Laboratory. The dataset includes two partitions: (1) the expanding shock-cylinder (CYL) simulations and (2) the Perturbed Layered Interface (PLI) simulations. Entries in both partitions consist of time series of arrays of thermodynamic fields (pressure, density, and temperature), kinematic fields (position and velocity), and additional fields that depend on thermodynamic and/or kinematic fields (e.g., material stress). Materials in the CYL partition include solids (aluminum, copper, depleted uranium, stainless steel, tantalum, and a generic polymer), a liquid (water), gases (air, nitrogen), and a generic detonating material. The PLI partition spans a highly varying geometry but consists of fixed materials across entries: copper, aluminum, stainless steel, polymer, and high-explosive. HEAT captures critical phenomena such as momentum transfer, shock propagation, plastic deformation, and thermal effects, making HEAT a valuable benchmark for development of AI/ML emulation of multi-material shock propagation.
\end{abstract}

\section{Value of the data}

\begin{itemize}
\item High explosives simulations are rich multi-physics models (e.g., detonation chemistry, fluid-structure interaction, and heat transfer), giving a more comprehensive understanding than what is possible with analytical formulas (which are typically single-physics) alone. Multi-physics means the coupling of various conservation laws with Equations of State for different materials and different models for different phases of the same material.
\item Eulerian constant-time-step, simulation-derived high explosive datasets are useful because they record the evolution of all the kinematic and thermodynamic variables on a regular grid at evenly spaced points in time. Experimental data of the same phenomena is sparser. Simulations can show internal jet formation, detonation wave propagation, and material deformation in detail that is otherwise impossible to see without ultra-high-speed imaging or X-rays.
\item Simulation-derived datasets reduce the need for experiments during early design stages of new high explosive configurations, allowing only the most promising designs to be tested physically.
\item These data are valuable because real-world high explosive testing is dangerous. High explosive configurations pose risks to personnel and equipment.
\item These data are valuable because of the high cost of real-world experiments. Experiments can be expensive due to the materials, safety precautions, and instrumentation required.
\item Training a predictive surrogate model on these data is a process that closely resembles video generation because these data are 2D with uniform spatial and temporal discretizations. Therefore, these data may be valuable for creative video generation.
\end{itemize}

\section{Background}

AI surrogate models of conventional multi-physics and multi-material models promise computationally efficient alternatives to conventional models with similar accuracy \cite{zhou2025predicting}. Here, a conventional model is a numerical model that solves discretized systems of partial differential equations that describe physical systems at continuum scales under various assumptions. Analytical solutions are generally infeasible except for limited cases \cite{evans2022pde}, so numerical methods that solve discretized equations with known convergence and accuracy properties are typically used. Typically, these numerical methods are computationally expensive when high accuracy is needed and computationally cheaper, lower resolution implementations tend to alias critical dynamics. The promise of fast yet accurate models motivate the development of AI surrogate models that capture only the essential features of conventional models at the relevant scale.
AI surrogate models of conventional multi-physics and multi-material models need diverse training data to forecast the evolution of coupled nonlinear dynamical systems \cite{ohana2024well}. Multi-physics AI surrogate models \cite{oberkampf2002verification} typically approximate solutions to autoregressive prediction problems in which model $f$ is trained to predict
\begin{align}
    U( x, t_{n+1} )=f( U(x, t_n ) ),
\end{align}
where $U$ is a state vector, $x$ are spatial coordinates, $t$ is time, and subscript $n$ is the time step index such that $t \in \{t_0,t_1,\dots,t_T\}$. AI surrogate model training data must be representative of many physical regimes to learn the functionality of conventional multi-physics models, to avoid overfitting, and to support generalization beyond the training data. To the author’s knowledge, no other publicly available dataset captures the interacting dynamics of myriad solids (aluminum, copper, depleted uranium, stainless steel, tantalum, and a generic polymer), liquids (water), gases (air, nitrogen), and detonator material (high explosive, HE) under shock loading and explosion. HEAT \cite{lanl2025oceans11} adds a rich menagerie of new physical phenomena to support the training of surrogate models.

\section{Data Description}

The PAGOSA hydrocode was used to generate a comprehensive suite of PLI simulations, consisting of roughly 5330 individual simulations each sampled at 101 evenly spaced timestamps (the first snapshot contains the initial conditions) for a total of 538,330 snapshots, and 2161 CYL simulations each sampled at 57 evenly spaced timestamps for a total of 123,177 snapshots. Thus, the HEAT dataset contains a combined total of 661,507 snapshots of PAGOSA simulation dynamics, with 7,491 of those snapshots the initial conditions. PAGOSA simulation units are centimeters, grams, microseconds, and Kelvin for length, mass, time, and temperature, respectively. The PLI and CYL simulations were computed with single precision.

\subsection{Perturbed Layer Interface (PLI) Simulations}

\begin{figure}
    \centering
\includegraphics[scale=0.75]{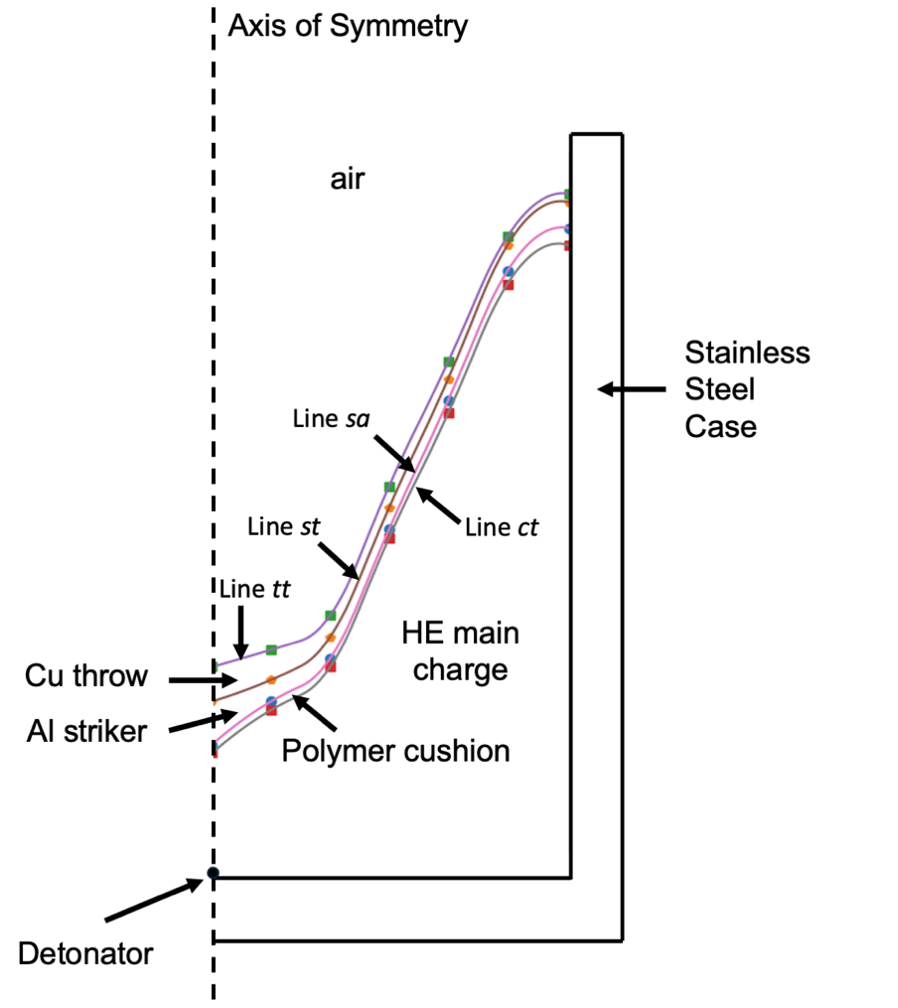}
    \caption{An example of the Perturbed Layered Interface at $t=0$. The 5330 simulations are composed of 5330 different liner geometries defined by the lines $ct$, $sa$, $st$, and $tt$, all other material and geometric parameters are held constant. Each simulation contains 100 different snapshots in time with an approximately $0.25 \, \mu s$ time step between them. At $t=0$ the detonator detonates and the HE burn begins.}
    \label{fig:pli_diagram}
\end{figure}

\begin{figure}
    \centering
\includegraphics[scale=0.75]{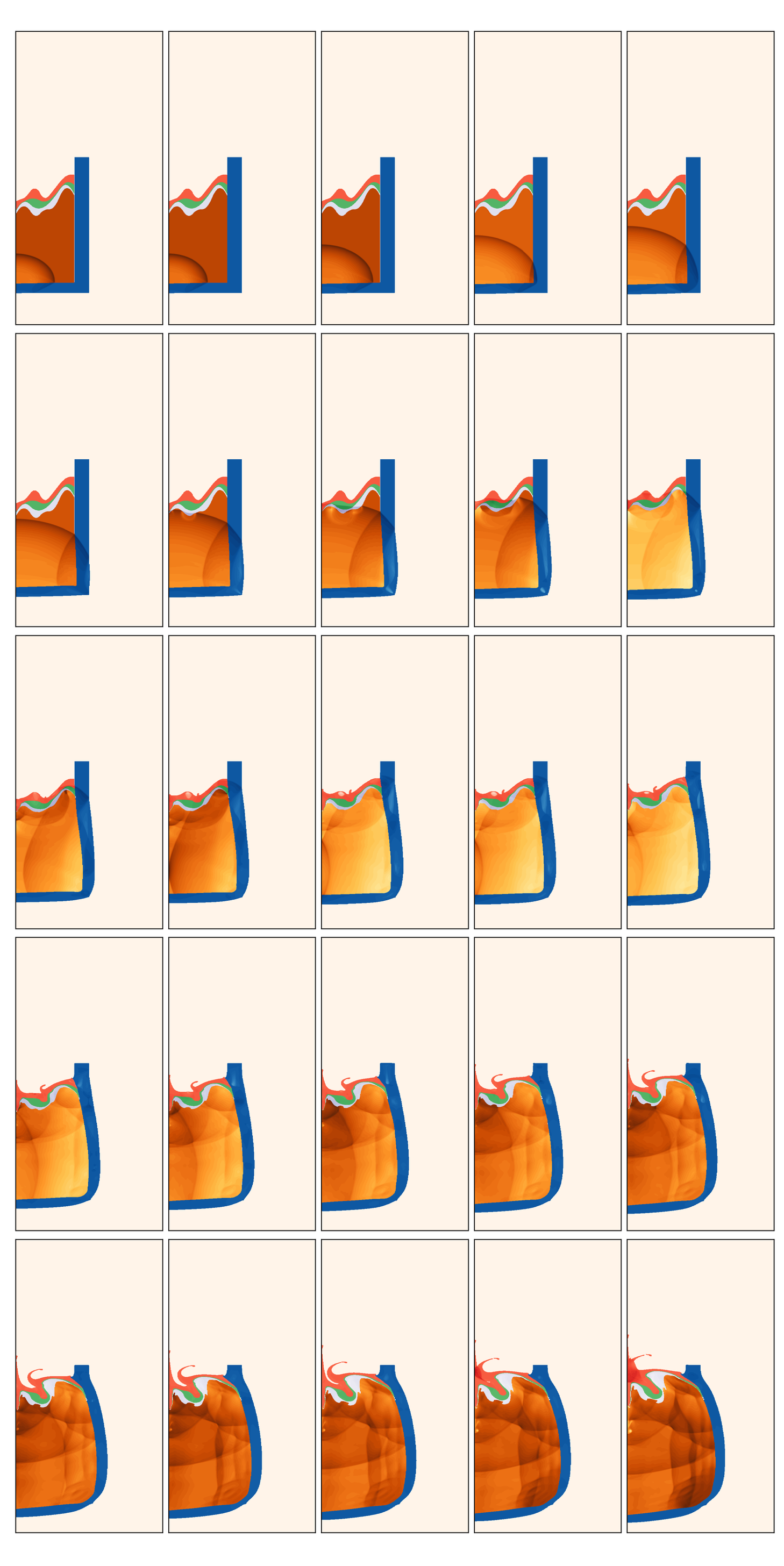}
    \caption{The evolution of PLI simulation \texttt{id00015}. Note the complex baroclinic jet geometry of the polymer cushion, Al striker, and Cu throw that forms around the axis of symmetry, the plastic deformation of the stainless steel case, and the reflected shock waves within the high explosive.}
    \label{fig:pli_dynamic}
\end{figure}

Each PLI simulation predicts the evolution of the high explosive (HE), stainless steel case, polymer cushion layer, Aluminum striker layer, Copper throw layer, and air in the configuration shown in Figure \ref{fig:pli_diagram} after the high explosive detonates at t = 0. Figure \ref{fig:pli_dynamic} shows the evolution of PLI simulation \texttt{id00015}. Time advances in increments of roughly 0.25 $\mu s$, reaching a final simulated time of $t = 25.0 \, \mu s$.  For every snapshot the code outputs a Eulerian grid of the full hydrodynamic field, including the $(R, Z)$ velocity components and the $(R, Z)$ spatial coordinates. In each grid cell the properties of every material present are recorded in numpy-zip (\texttt{*.npz}) file arrays for every time step and every material: density, temperature, pressure, volume fraction, stress, strain, shear modulus, yield strength, plastic strain (PLST), cell‑average yield (Yield), and sound speed. The only degrees of freedom that differ from run to run are the six spline coefficients that define the four interface‑shape spline functions: $sa$ (striker anchor), $st$ (striker thickness), $ct$ (cushion thickness), and $tt$ (throw thickness). These four lines define the interfaces between the polymer cushion and aluminum striker, between the aluminum striker and the copper throw, between the high explosive and the polymer cushion, and between the copper throw and air, respectively.

The coefficients of the spline functions for each simulation at $t=0$ are listed in the catalogue file, \texttt{pli.csv} hosted in the directory \texttt{/heat/pli/} on \texttt{oceans11.lanl.gov} \cite{lanl2025oceans11}. The simulations (listed as \texttt{id00001}, \texttt{id00002}, etc.) can be downloaded individually in \texttt{/heat/pli/pli240420}. Within each simulation directory individual time steps may be downloaded. For example, \texttt{/heat/pli/pli240420/id00001} contains 101 individual numpy-zip files that each contain the hydrodynamic data at one snapshot in time. The entire dataset may be downloaded at once by downloading \texttt{/heat/pli/pli240420.tar}. The data sets on \texttt{oceans11.lanl.gov} with suffix \textbf{\_full} and \textbf{\_half} refers to the spatial resolution, where the \textbf{\_half} was uniformly downsampled to have half the spatial resolution of \textbf{\_full}.

\subsection{Expanding Shock-Cylinder (CYL) Simulations}

\begin{figure}
    \centering
\includegraphics[scale=0.75]{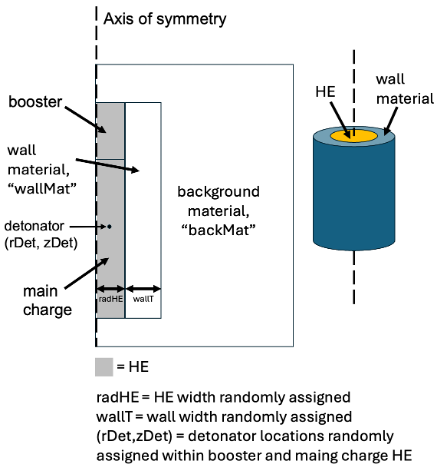}
    \caption{An example of the Expanding Shock-Cylinder at $t=0$. The detonation location (\texttt{rDet}, \texttt{zDet}), HE width (\texttt{radHE}), cylinder wall thickness (\texttt{wallT}), wall material, and background material are all varied. Each simulation contains 57 different snapshots in time with an approximately $0.25 \, \mu s$ time step between them. At $t=0$ the detonator detonates and the HE burn begins.}
    \label{fig:cylex_diagram}
\end{figure}

\begin{figure}
    \centering
\includegraphics[scale=0.75]{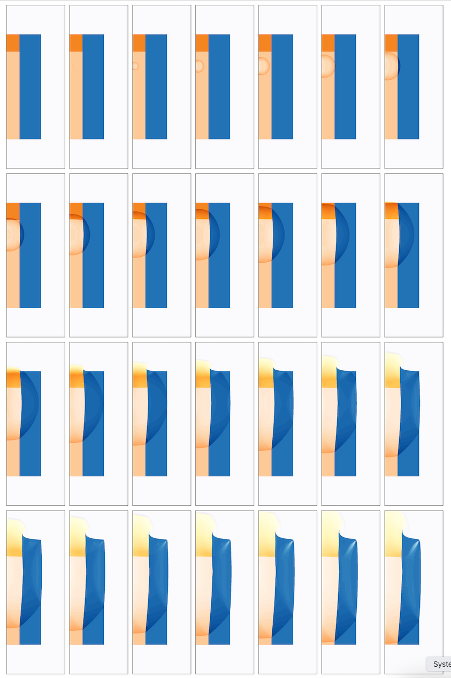}
    \caption{The evolution of CYL simulation \texttt{id00433}. The detonation in the upper part of the high explosive near the axis of symmetry causes the high explosive booster to move outward into the air background, while shock waves within the aluminum cylinder wall create plastic deformation along the upper surface of the aluminum cylinder.}
    \label{fig:cylex_dynamic}
\end{figure}

Each simulation in the cylex (CYL) collection represents a two-dimensional cylindrically symmetric configuration: a tube of high explosive (HE) is encased by a cylindrical wall, which in turn is surrounded by a background material (Figure \ref{fig:cylex_diagram}). Figure \ref{fig:cylex_dynamic} shows the evolution of CYL simulation \texttt{id00433}, in which the high explosive is surrounded by an aluminum wall and air background. The thickness of the HE, the thickness of the cylinder surrounding the HE, and the detonator location are all randomly varied. The energy source consists of two HMX derived HE components (a booster and a main charge) with prescribed burn rates. The wall and background materials are varied for nearly all possible combinations of materials, drawing from six strength modeled materials (Cu, Stainless Steel, Al, Be, Depleted Uranium, Ta) and four strength-less media (air, water, nitrogen, polymer) in their equilibrium states. Physical processes captured include momentum transfer, HE‑driven momentum divergence/energy input, heat transfer, shock propagation, compressible flow, and plastic flow.

For every CYL snapshot (numpy-zip file) the following fields are stored: density, pressure, internal energy, and volume‑fraction fields for each material (3–4 material density fields per simulation), velocity and position fields. Within \texttt{oceans11.lanl.gov/heat/cyl} is \texttt{cyl.csv}, which contains the key necessary for understanding the arrays in the numpy-zip files for each CYL simulation. For example, \texttt{cyl.csv} shows that the simulation \texttt{cx241202\_id00320} was of a cylinder made of aluminum (\texttt{wallMat}) surrounded by tantalum (\texttt{backMat}). The high explosive detonator was placed at the $(r, z)$ (\texttt{rDet} and \texttt{zDet}) location 0.211, -2.889 and the radius of the high explosive (\texttt{radHE}) was 0.665 (cm) while the aluminum cylinder wall thickness (\texttt{wallT}) was 4.608 (cm). The simulations (listed as \texttt{id00001}, \texttt{id00002}, etc.) can be downloaded individually in \texttt{/heat/cyl/cx241202}. Within each simulation directory individual time steps may be downloaded. For example, \texttt{/heat/cyl/cx241202/id00001} contains 57 individual numpy-zip files that each contain the hydrodynamic data at one snapshot in time. The entire dataset may be downloaded at once by downloading \texttt{/heat/cyl/cx241202.tar}. The data sets on \texttt{oceans11.lanl.gov} with suffix \texttt{\_full} and \texttt{\_half} refers to the spatial resolution, where the \texttt{\_half} was uniformly downsampled to have half the spatial resolution of \texttt{\_full}.

\section{Experimental Design, Materials, and Methods}

Both the CYL and PLI sub-datasets were generated using the PAGOSA code of Los Alamos National Laboratory (LANL). PAGOSA is a computational fluid dynamics code developed for simulating high-speed compressible flows and high-rate material deformation \cite{culp2025pagosa}. It is a Eulerian reference frame finite-difference code designed to handle complex interactions among gases, fluids, and solids, with broad support for diverse equations of state (EOS), material strength models, and explosive behavior.

PAGOSA is particularly suited for studying extreme physical phenomena such as explosively driven systems, high-velocity impacts, and other high-pressure, high strain-rate events, where pressures range from kilobars to megabars. In this regime, materials undergo significant volume changes, making incompressibility assumptions invalid. The code is intended to resolve the propagation of compression and rarefaction waves within materials—critical for accurately modeling dynamic events in continuum mechanics. Accordingly, PAGOSA has been used to simulate shaped charges \cite{miers2025shapedcharge}, explosively fired projectiles \cite{subramanian2024codetocode}, and other high explosive tests for proton radiography \cite{smith2024boron}. 
High explosive detonation was model using the FSD (Fast Sweeping Detonation) burn model is an alternative programmed burn method which calculates burn times by solving the Eikonal equation using the Fast Sweeping Method \cite{culp2025pagosa}. The FSD burn model was computed with second order spatial accuracy.

The material strength model is the Standard Strength Model. The Standard Strength Model is an isotropic strength model and can use any of the flow-stress models (also known as yield strength), which are discussed below. Also, this model can use any one of the fracture models listed below in this manual. This is an isostropic, elastoplastic strength model with radial return of the yield surface. By default, it uses the Jaumann stress rate that is objective with respect to rigid body rotation and translation. A material is allowed to undergo an elastoplastic transition. The yield criterion determines the transitional point at which material stops deforming elastically and begins to deform plastically. The model uses the von Mises (radial return) criteria to get that point. Then the stress deviators are adjusted so they will remain on the yield surface during plastic flow. The flow-stress model calculates the yield strength and shear modulus. No damage models were used to model material fracture.

After a material can no longer sustain any strength (generally due to a change of phase to continuously deformable fluid). In this case, PAGOSA solves a discrete form of the hydrodynamic equations \cite{culp2025pagosa}:
\begin{align}
\frac{\partial \rho}{\partial t}
+ u \frac{\partial \rho}{\partial x}
+ v \frac{\partial \rho}{\partial y}
&= -\rho \left( \frac{\partial u}{\partial x}
+ \frac{\partial v}{\partial y} \right), \\[6pt]
\frac{\partial u}{\partial t}
+ u \frac{\partial u}{\partial x}
+ v \frac{\partial u}{\partial y}
&= -\frac{1}{\rho} \frac{\partial p}{\partial x}
+ \frac{F_x}{\rho}
+ \frac{1}{\rho} \left(
\frac{\partial S_{xx}}{\partial x}
+ \frac{\partial S_{xy}}{\partial y}
\right), \\[6pt]
\frac{\partial v}{\partial t}
+ u \frac{\partial v}{\partial x}
+ v \frac{\partial v}{\partial y}
&= -\frac{1}{\rho} \frac{\partial p}{\partial y}
+ \frac{F_y}{\rho}
+ \frac{1}{\rho} \left(
\frac{\partial S_{yx}}{\partial x}
+ \frac{\partial S_{yy}}{\partial y}
\right), \\[6pt]
\frac{\partial e}{\partial t}
+ u \frac{\partial e}{\partial x}
+ v \frac{\partial e}{\partial y}
&= -\frac{p}{\rho} \left(
\frac{\partial u}{\partial x}
+ \frac{\partial v}{\partial y}
\right)
+ \frac{1}{\rho} \left(
S_{xx} e_{xx}
+ S_{yy} e_{yy}
+ 2 S_{xy} e_{xy}
\right).
\end{align}
where the strain rate tensor elements are
\begin{align}
e_{xx} &= \frac{\partial u}{\partial x}, \\[6pt]
e_{yy} &= \frac{\partial v}{\partial y}, \\[6pt]
e_{xy} &= \frac{1}{2} \left(
\frac{\partial u}{\partial y}
+ \frac{\partial v}{\partial x}
\right).
\end{align}
and the deviatoric stress tensor elements are
\begin{align}
S_{xx} &= 2G \left(
e_{xx} - \frac{1}{3} \left(
\frac{\partial u}{\partial x}
+ \frac{\partial v}{\partial y}
\right)
\right), \\[6pt]
S_{yy} &= 2G \left(
e_{yy} - \frac{1}{3} \left(
\frac{\partial u}{\partial x}
+ \frac{\partial v}{\partial y}
\right)
\right), \\[6pt]
S_{xy} &= 2G\, e_{xy}.
\end{align}
where $G$ is the shear modulus. An artificial viscosity model is required to stabilize the simulations and accurately capture shock waves without introducing nonphysical oscillations. We use the Wilkins model \cite{culp2025pagosa} for artificial viscosity.

The EOS specifies the pressure for a given material as a function of the density and specific internal energy. The EOS tables for each material are those described in \cite{sheppard2025sesame}.

\section{Software for Training on HEAT}

The HEAT dataset consists of numpy-zip files and is therefore easily readable through numpy load facilities. Once a numpy-zip is loaded the python environment has a dictionary of arrays representing the various hydrodynamic fields in the PLI or CYL simulations. Names of these fields make up the keys of the dictionary and corresponding numpy arrays for each field make up the values of the python dictionary.

For training large neural network models on this data we provide a readily installable python module called \href{https://github.com/lanl/Yoke}{Yoke} \cite{lanl2025yoke} which includes pytorch dataset classes which create various input-output pairs of pytorch tensors. In particular, the autoregressive training task associated with the PLI and CYL raw data are defined in the classes \texttt{LSC\_rho2rho\_temporal\_DataSet} and \texttt{TemporalDataSet}, respectively. These classes, along with other helpful utilities, are located in the modules \url{https://github.com/lanl/Yoke/blob/main/src/yoke/datasets/lsc_dataset.py} for PLI and \url{https://github.com/lanl/Yoke/blob/main/src/yoke/datasets/load_npz_dataset.py} for CYL.

\section{Limitations}

None of the materials in the PLI and CYL data included a damage model for solids subjected to shock loading. The simulated solids can only deform elastically and plastically and never break, crack, or spall, where spall is defined as tensile fractures that arise from reflected shock wave patterns within a solid. Therefore, the solid dynamics captured in HEAT over-predict load-carrying capacity and survivability because shocks are absorbed that may otherwise cause fractures and spallation. This shock absorption can cause secondary inaccuracies due to non-physical temperature increases, plastic work, and wave damping.

Conventional models that use a Eulerian hydrodynamic reference frame, such as PAGOSA, model fluid motion by solving the governing equations on a fixed spatial grid through which the fluid flows. These discrete models have many sources and types of errors \cite{oberkampf2002verification}. For example, as the materials move through the grid sharp interfaces (e.g., between metal and air) can become blurred over time due to numerical diffusion. This can cause non-physical states, such as metal-air interface cells with unrealistic densities or pressures. Flux calculation numerical errors can lead to small violations of conservation laws that steadily accumulate as the forecast progresses in time. Interface tracking numerical errors that occur in mixed-material cells can also cause cumulative errors in global quantities and non-physical heating or mass loss/gain. Fixed spatial grids can also impose artificial directional bias. Finally, different materials have different Equations of State (EOS) that are sources of epistemic error \cite{oberkampf2002verification}, which arise from a lack of understanding of the underlying physical processes. Weighted averages of EOS values in interface cells can create thermodynamic inconsistencies. Long term forecasts of coupled and highly nonlinear systems generate aleatoric errors \cite{oberkampf2002verification}, which arise from intrinsic randomness. Together, these factors highlight how both epistemic and aleatoric uncertainties, compounded by numerical and modeling errors inherent to Eulerian formulations, can significantly degrade the physical fidelity and predictive reliability of conventional hydrodynamic simulations.

\section{CRediT Author Statement}

\textbf{Kyle Hickmann}: Conceptualization, Software, Methodology, Editing, Supervision. \textbf{Bryan Kaiser}: Data Curation, Writing, Original draft preparation, Editing, Software, Methodology. \textbf{Divya Banesh}: Data Curation. \textbf{Jesus Pulido}: Data Curation. \textbf{Sharmistha Chakrabarti}: Software, Methodology. Christine Sweeney: Supervision, Data Curation. \textbf{Soumi De}: Software, Methodology. \textbf{Sourabh Pandit}: Data Curation, Software. \textbf{David Schodt}: Software.

\section{Acknowledgements}

This research did not receive any specific grant from funding agencies in the public, commercial, or not-for-profit sectors. This work was supported by the U.S. Department of Energy through the Los Alamos National Laboratory. Los Alamos National Laboratory is operated by Triad National Security, LLC, for the National Nuclear Security Administration of U.S. Department of Energy (Contract No. 89233218CNA000001). This manuscript is cleared for unlimited release by the U.S. Department of War (DOPSR release number 26-T-0965) and by Los Alamos National Laboratory (LA-UR-26-22536). This clearance does not imply U.S. Department of War endorsement or factual accuracy of the material. The data presented here (available at oceans11.lanl.gov) is also approved for unlimited release (LA-UR-25-26490).

\bibliography{main}
\bibliographystyle{iclr_templates/iclr2025_conference}

\end{document}

%% file: iclr_templates/math_commands.tex
\usepackage{amsmath,amsfonts,bm}
\usepackage{booktabs}

\def\eqref#1{equation~\ref{#1}}

\def\1{\bm{1}}

\DeclareMathAlphabet{\mathsfit}{\encodingdefault}{\sfdefault}{m}{sl}
\SetMathAlphabet{\mathsfit}{bold}{\encodingdefault}{\sfdefault}{bx}{n}

